# Movement Control of Smart Mosque's Domes using CSRNet and Fuzzy Logic Techniques

Anas H. Blasi[1], Mohammad Awis Al Lababede[2], Mohammed A. Alsuwaiket[3]
Computer Information Systems Department, Mutah University, Al Karak, Jordan[1]
Computer Science Department, Mutah University, Al Karak, Jordan[2]
Computer Science and Engineering Technology Department, Hafar Batin University, Hafar Batin, Saudi Arabia[3]

*Abstract*—Mosques are worship places of Allah and must be preserved clean, immaculate, provide all the comforts of the worshippers in them. The prophet's mosque in Medina/ Saudi Arabia is one of the most important mosques for Muslims. It occupies second place after the sacred mosque in Mecca/ Saudi Arabia, which is in constant overcrowding by all Muslims to visit the prophet Mohammad's tomb. This paper aims to propose a smart dome model to preserve the fresh air and allow the sunlight to enter the mosque using artificial intelligence techniques. The proposed model controls domes movements based on the weather conditions and the overcrowding rates in the mosque. The data have been collected from two different resources, the first one from the database of Saudi Arabia weather's history, and the other from Shanghai Technology Database. Congested Scene Recognition Network (CSRNet) and Fuzzy techniques have applied using Python programming language to control the domes to be opened and closed for a specific time to renew the air inside the mosque. Also, this model consists of several parts that are connected for controlling the mechanism of opening/closing domes according to weather data and the situation of crowding in the mosque. Finally, the main goal of this paper has been achieved, and the proposed model has worked efficiently and specifies the exact duration time to keep the domes open automatically for a few minutes for each hour head.

## I. INTRODUCTION

Islam is the second largest religion after Christianity in the world, according to a study conducted in 2015 [1], Islam has 1.9 million followers in the world, representing 24.8% of the world's population.

Modern technology has become an influencing factor in our lives and has reached a stage that we cannot do without because it has become a common factor in the affairs of our lives in general. Among the most important aspects of the spread of technology in the modern era are techniques of artificial intelligence that stimulate the process of automating all areas of life to make it easier, faster, and more intelligent in adapting to the tremendous development that we are witnessing from time to time, as it has become an essential element in the field of learning, working, providing services, agriculture, trade, industry, and many others, which helped to improve results desired and achieving great benefits for employers with less time, effort and cost.

Here comes our role as scientists to support the techniques of artificial intelligence to employ them as much as possible, so this paper considered using these technologies to solve the problem of overcrowding in mosques. The Islamic religion is considered the second most widespread religion in the world after the Christianity religion [1], and it is distinguished by special rites distinct from other religions, the most important of which is prayer, especially collective prayer in mosques, which means they need for mosques to spread in line with the large numbers of worshipers and provide a comfortable and clean place to perform the prayer Easily and conveniently. There are four million mosques scattered around the world [2] and it is equipped with the best hygiene and health protocols and this is what makes it one of the cleanest places in the world to perform Muslim prayers. Among the most important of these mosques:

- Al Masjid Al Haram [3]: It has high importance and priority for Muslims, located in Makah Al-Mukarramah/ Kingdom of Saudi Arabia, it is the qibla of the Muslims and there is the honorable Kaaba which Muslims visiting it from all parts of the world to perform the Hajj and Umrah and one prayer in it equal to one hundred thousand prayers in other mosques.

- The Prophet's Noble Mosque [3]: No less important than its predecessor, where the Messenger of the Nation of Islam, Muhammad (peace and blessings of God be upon him) lived and was buried there with two of his companions (Omar and Abu Bakr - may God be pleased with them both) - located in Medina / Kingdom of Saudi Arabia, it is frequented by Muslims from all over the world to visit the tomb of the Prophet and the prayer in it is equivalent to 1000 prayers in others.

In this paper, the focus will be on the Prophet's Noble Mosque as a model that can be converted into a smart mosque, being one of the largest mosques in the world and the second holiest site in Islam after Al-Masjid Al-Haram in Mecca, which is the mosque that the Prophet Muhammad built in Medina after his migration in 1 Hijra next to his house [4]. The mosque has gone through several expansions throughout history, as it now covers 98,327 square meters for the building and 235,000 square meters for the surrounding squares [4], which can accommodate approximately 698,000 worshipers, and its expansion is ongoing to reach a capacity of 2.5 million worshipers [5]. The mosque building has many domes: the green dome (main dome), 170 gray dome, and 27 moving domes, in addition to 10 minarets.

As noted above, the mosque accommodates a huge number of worshipers, this means that an appropriate environment of moderate temperature and humidity must be provided to help prevent the spread of some viruses such as Coronavirus, especially that the mosque contains carpets, which is an appropriate environment for their transmit, which may cause the transmission of infections to worshipers, and also it can be noted that the mosque contains 27 moving domes, so it is possible to use these domes to achieve the goal of this paper, which is to improve the quality of air and humidity in the Holy Prophet's Mosque by controlling the opening of domes to replenish the air and the sunlight entering the mosque building using Congested Scene Recognition Network (CSRNet) and Fuzzy techniques based on temperature and the number of worshipers in the mosque without the need for human intervention in this process.

The idea came to address a problem in previous research [6], as domes were opened depending on some weather factors, but a problem with temperature was observed so that domes did not open unless the temperature was between (16°-27°), and given the extremely volatile climate in Medina, in case of high temperatures (more than 27°), the domes will remain closed even if the mosque needs ventilation, and also in very cold weather (less than 16°), it will remain closed, but in this paper, the number of worshipers was inserted to improve the work of domes in the required ventilation.

The paper has organized as follows: Section II reviews the related work. Section III describes the materials and methods used to build the proposed model. Section IV discusses the results in detail. Finally, Section V discusses the conclusions and draws the future work.

## II. Related Work

As mentioned earlier, this research is an extension of a previously published paper [6] that aims to solve the problem of ventilation, allow sunlight to enter the mosque, and provide comfort for the worshipers, especially in times of congestion in mosques and the solution was by building a model for smart mosque domes using weather features and outside temperatures. Machine learning algorithms such as k-Nearest Neighbors (kNN) and Decision Tree (DT) were applied to predict the state of domes (open or close). The experiments of this paper were applied to the Prophet's Mosque in Saudi Arabia, which mainly contains twenty-seven hand-moved domes. Each of the machine learning algorithms was tested and evaluated using different evaluation methods. After comparing the results of both algorithms, the DT algorithm achieved 98% higher accuracy compared to 95% accuracy for the k-NN algorithm. The problem with this paper was in the element of temperature, as the domes do not open except in the area (16°-27°), even if the mosque needs ventilation.

In [7] a neural network has been proposed to discover crowded scenes called (CSRnet) as it is one of the deep learning methods that can understand the very crowded scenes and make an accurate estimation of the count, consisting of two layers: a neural network as a front face and an expanded network for the back end, the authors made experimentations on four groups of images: (Shanghai Tech, WorldExpo 10, UCFCC50, UCSD), the results were so high that they reduced the error by 47.3% while improving by 15.4% compared to the previous methods that were applied to the same group of images.

The author in [8] presented a conceptual model for heating and controlling air conditioning inside the home by applying the principle of fuzzy logic, microprocessors associated with sensors were used to sense the factors affecting ventilation in addition to a compressor and an air circulation fan, all of which were installed inside a building for testing. The proposed model aims to provide comfort and energy-saving by regulating the airflow to different areas of the house depending on the ambient and external temperatures in addition to the relative humidity as parameters according to the rule base, so that the outputs are a compressor speed setting (increase or decrease), adjust the fan speed (Increase or decrease), change the mode of the air conditioner to (hot, cold / off), open or close the ventilation zone. Based on the analysis of the data collected and tested, a comfortable atmosphere was obtained throughout the house with the belief that energy use is very efficient.

The author in [9] is an application of the Fuzzy logic in the field of agriculture and ventilation of greenhouses, the aim of which is to set the appropriate atmosphere for plant growth within greenhouses based on the close relationship between temperature and humidity using a physical model that works to decouple this relation in-between them to manage the internal climate while saving energy as a result of reducing When the engine is running, this system was simulated using a MATLAB so that the temperature was set to 20 Celsius at night and 28 Celsius in the morning with the adopted relative humidity (70%) to complete the ventilation process through entering the more humid air into the greenhouses to maintain the degree of Relative humidity, the study results were similar to the expected results, but more study is needed in detail during operation to verify energy saving.

The authors presented in [10] a prototype of the crowd estimation system in Al-masjid Al-Haram, which helps in guiding the mosque's visitors and managing the crowds in the place through visual representation in the form of a thermal map (i.e., crowd representation as a thermal block), this work was divided into two parts:

- Explore the system's features, aspects, and design stages.

- Preparing a short comparison between two textile-based techniques used to estimate the crowd: LPB & GLCM.

The data used was collected through the mosque's cameras (Sa'i area between Safa and Marwa in particular) to analyze it to estimate the crowd density, then converted it into thermal maps to know the crowded places in the mosque so that it provides the opportunity for users to avoid these crowds or redirect to less crowded areas, for the results, The LPB algorithm yielded comprehensive results of 87.9%, especially for the top and side images.

The [11] aims to develop a logic-based smart ventilation system to control indoor air quality in pharmaceutical sites because indoor air quality affects the pharmaceutical industry, production, and storage, including appropriate temperature, humidity, airflow, and the number of appropriate microorganisms, the proposed system works depending on the fuzzy inference system, where the ventilation system can control airflow and quality according to internal temperature, humidity, airflow and microorganisms in the air. The MATLAB Fuzzy Logic Toolbox was used to simulate the performance of the fuzzy inference system. The results showed that the temperature difference has less effect on controlling the position of the system but has a noticeable effect on air conditioning and fan speed. If the temperature difference corresponds to the "hot" organic function, the air conditioning and fan speed are set to the "medium" organic function, and the higher the temperature difference from "warm" to "hot", the air conditioning and the fan will increase the speed to "very fast" and the system efficiency can be improved by processing input and output parameters according to user requirements.

In fuzzy ventilation control for zone temperature and relative humidity [12] first goal is to use free cooling and drying of available humidity due to the differences in the area and surrounding conditions and this is done by changing the ratio of fresh air that enters the heating, ventilation and air conditioning system and then the controlled area, while the other goal is to maintain the conditions of the region at a preferred control point located between the top and bottom turning points so that the upper and lower turning point boundaries and the preferred set point are adjusted for fuzzy ventilation control purposes to ensure occupant comfort. The HVAC plant becomes active when the fuzzy ventilation control strategy is unable to keep area conditions within the maximum and minimum points of the point. Simulation results were compared using a fuzzy ventilation control strategy with normal plant operation using PID controllers. This standard comparison was used to evaluate the benefits of using a fuzzy ventilation control strategy. The comparisons lasted for 52 weeks based on certain weather data that make sure of exposure to various ambient weather conditions, which result in it the ability of the HVAC station to operate between 05:00 and 17:00 daily.

The author in [13] proposes a physical model of greenhouse used in the Simulink / MATLAB environment to simulate the internal temperature and humidity as a mechanism for controlling air temperature and humidity in greenhouses. A fuzzy logic method was developed to control motors that are installed inside the greenhouse for heating, ventilation, humidification, and cooling to obtain a suitable local climate, the results showed a stable behavior of both temperature and humidity with a low rate of heating, ventilation, and humidification without the need to use a dehumidifier system to reduce energy consumption.

In [14], a hybrid learning algorithm is proposed that represents a general model of the neural network for controlling fuzzy logic and decision systems. This model combines the idea of controlling fuzzy logic, the structure of the neural network, and learning capabilities in an integrated mysterious logical control system based on the neural network and the decision-making system especially in the speed of learning.

The author in [15] suggested a self-adjusting Fuzzy PI controller in the HVAC air pressure control loop. Fuzzy Self-Adjusting Console (STFPIC) online output measurement factor is modified by vague rules according to the current direction of the controlled process. The base rule is defined to set the resulting measurement factor to error and change the error of the controlled variable. Ziegler-Nichols PI tuner or PID controller works well around normal working conditions but tolerating it to processing parameter changes is highly affected. STFPIC has been used to overcome these shortcomings. Compared with the PID and Adaptive Neural Controls (ANF), simulation results show that STFPIC performance is better under normal conditions as well as when the HVAC system encounters major differences in parameters.

The [16] analyzed the dynamics of the crowd for visitors at the Prophet's Mosque during the most saturated period to describe the most dangerous conditions and suggest technical solutions to accommodate visitors and provide them with a safe passage. The main purpose of the statistical analysis in this study is to investigate the current numbers of visitors to the Prophet's Mosque and prepare administrative plans for future expansion to accommodate the expected number of visitors within specific sites in the mosque. Data was collected by performing an actual count of visitors from the videos and recorded images taken by the legal authority during the holy month of Ramadan and the month of Dhu al-Hijjah and during the busiest hours of the day such as the time of entry from the Peace Gate to the Prophet Mohammed's tomb and from the tomb to the exit from the Baqi Gate. This study provides reasonable information on the crowd dynamics that can be adopted by any responsible crowd management authority aiming to accommodate a large number of visitors during the busiest seasons without causing any harm to visitors. The results of this study are expected to help improve crowd management in the mosque.

### III. Methodology

In this section, the methods and materials will be described and discussed through showing the implementation of the counting people technique using the Congested Scene Recognition Network (CSRNet) algorithm and build the fuzzy control system by using the data weather [17], also an example for the whole system will be proposed.

This system consists of several parts that connected for controlling the mechanism of opening/closing domes according to weather data and the situation of crowding in the mosque, in general, the following tools and techniques are all we need to build the model that have proposed:

- Camera: First, a camera with high specifications and quality is needed to take pictures from all parts of the mosque to later analyze these images using the algorithm CSRNet and get the desired result which is the approximate number of worshippers, which is available in the Prophet's Mosque and can be used.

- Rainfall Sensor 3864: It should be placed at the top of the dome to predict precipitation accurately and reliably, this type is lightweight, frost-proof, and heat resistant [18], which is good to detect rainfall in real-time and force domes to close.

- DHT22 SENSOR: It is used for measuring temperature and humidity [19]. It uses a capacitive humidity sensor and a thermostat to measure the surrounding air. This sensor is cost-effective, provides low power consumption, good for -40 to 80°C temperature readings with ±0.5°C accuracy, and up-to 20meter signal transmission is possible.

- ARDUINO UNO: The microcontroller used here is an Arduino UNO [20]. The UNO is a microcontroller board based on ATMEGA 328P. The ATMEGA 328P has 32kB of flash memory for storing code. The board has 14 digital input and output pins, 6 analog inputs, 16 MHz quartz crystal, USB, an ICSP circuit, and a reset button. The UNO can be programmed with the Arduino software.

- Steel Rails: It should have the ability to withstand friction caused by moving the domes, where domes are placed on these tracks to move in one direction to be opened/closed (Fig. 1).

The images have taken from all over the mosque through the cameras distributed inside it. Then these images enter in the worshiper numbers predicting system using the captured images (CSRNet) techniques are applied, and then the congestion percentage in the mosque is calculated by dividing the number of worshipers on the capacity of the mosque. At the same time, rainfall and weather statuses are detected, and if there is rain, the domes are closed, else the weather statuses are entered along with the congestion rate into the fuzzy control system and the rules and relationships between the different data are built and then the dome is opened for a specific period, based congestion rate and weather factors.

*A. Data Set*

For the data of this research, it consists of two groups as follows:

*1) Weather data set:* It represents information about the weather conditions in the Kingdom of Saudi Arabia, obtained from the Kaggle.com website [17], represented by 249024 rows and 15 columns that is contained data for all the cities of Saudi Arabia, the hourly changing weather from 2017 to 2019, the columns are: date, hour, minute, day, temperature, humidity, wind strength, barometer, and visibility.

*2) A set of pictures:* to predict the number of worshipers in the mosque: It is a group of pictures that are taken and analysed through a CSRNet algorithm to predict the approximate number of worshipers at the mosque. A group of images was obtained from the Shanghai Technology Database [21] to train the model on it.

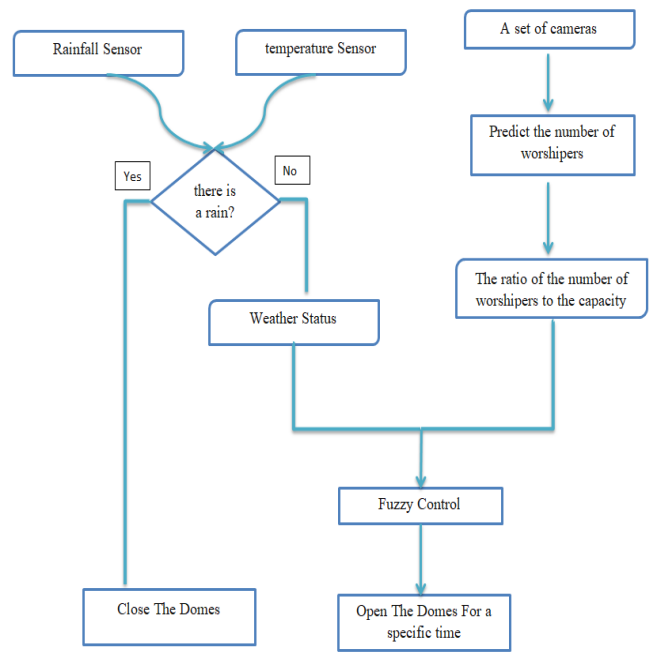

Fig. 1. Proposed Model of Fuzzy Control.

*B. Building the CSRNET Model*

Convolutional Neural Network (CNN) is one of the deep learning techniques and one of the types of the feed-forward neural network, depends on simulating the biological processes occurring in the visual lobe in the brain of living organisms and is used to solve computer vision problems in artificial intelligence and digital image processing and consists of an input layer, hidden layers, and an output layer [22], [23], [24], and to find the optimal solution and larger values of the real results the Back propagation technology is used in calculating the error rate each time and trying to reduce it.

To recognize the number of people in the images, the python programming language will be used to build the crowding detection algorithms using CSRNet algorithm. Where the proposed model firstly predicts the number of worshipers in the mosque by analyzing the group of images that will be taken hourly for the different parts of the mosque and then calculating the approximate total number of worshipers using CSRNet algorithm.

CSRNet algorithm is a technique for crowd counting by estimating the number of people in an image. It is Convolution Neural Networks (CNN) based methods, which are building an end-to-end recession method using CNNs Instead of looking at the spots of an image. Furthermore, it takes the entire image as input and directly outputs the number of crowds. CSRNet publishes a deeper CNN to catch high-level features and produce high-quality density maps without expanding the network complexity. CSRNet uses VGG-16 at the front end because of its strong transfer learning ability [7]. The output size from VGG is ⅛ of the original input size. CSRNet also uses dilated Convolution layers in the back end.

*C. Ground Truth Preprocessing*

In this part, to achieve the pre-processing step some important libraries like (torch, SciPy) will be used. Also, to build a two-dimensional density map for ground truth, the Gaussian kernel will be used to compute the real values of the people in each image in the dataset [21] bypassing the values, where the dataset contain the values, and these values consisting of the head annotations in that image, and the Gaussian filter and tree with k= 4 will be used.

*D. Implementation and Results of CSRnet Algorithm*

GPU was used to build the model by using the Cuda function in Python and to build the CSRNet model like [7], the straightforward way and end-to-end structure will be used. Also, a VGG-16 network with only use $3 \times 3$ kernels will be used for the front-end. For training phase. The accurate weights trained by the authors in [7] will be applied, and to show the accuracy of the results, it will be compared with the ground truth. CSRNet algorithm was tested on the Shanghai dataset [21] and get the results shown in Fig. 2.

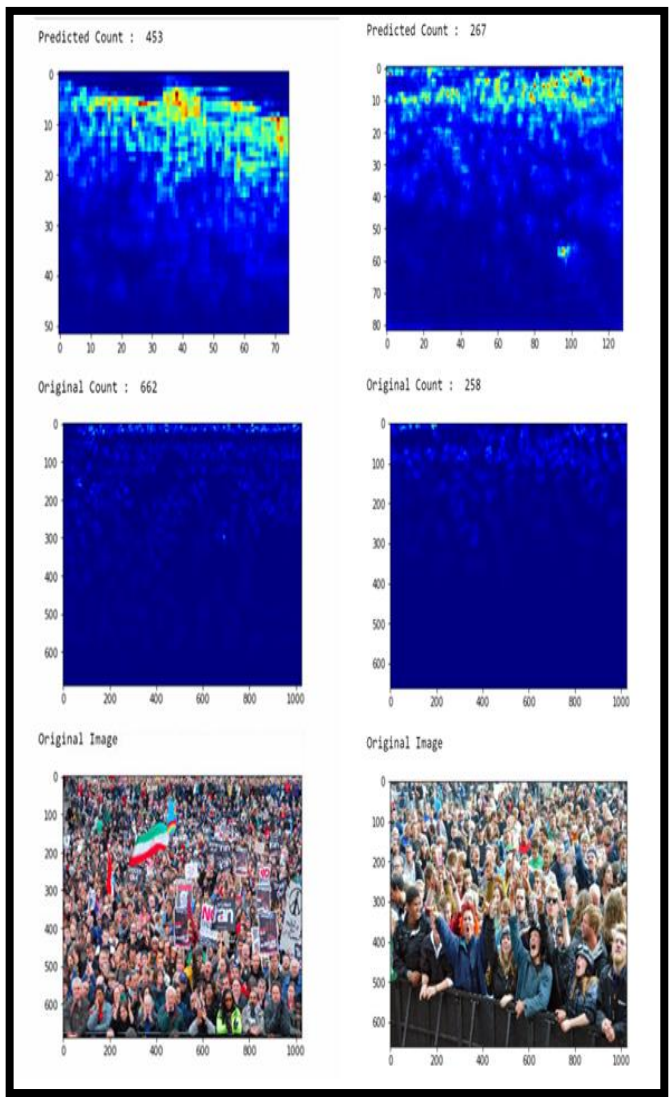

Fig. 2. Result of Counting People using CSRnet Algorithm.

Due to the difficulty of obtaining the number of worshippers inside the mosque, the algorithm was applied on the Shanghai Tech dataset (the same mechanism for any images). In Fig. 2, the image on the right shows that the actual number of people is 258, and after applying the CSRNet algorithm with the weights shown previously [7], an approximate number of 267 people was achieved. For the second image, the number of people in the original image is 662 and while an approximate number is 453 was achieved. Since an estimated percentage number of worshipers is needed in the mosque, it does not matter if it is 100% accurate for the proposed system. The previous results can be adopted, and the number of worshipers can be converted to a percentage.

*E. Moving the Domes using Fuzzy Control*

Fuzzy logic is a form of knowledge representation used in areas where concepts are difficult to define precisely, and which depend on their context for their understanding [25], where the Fuzzy Logic was introduced by Lotfi Zadeh and Berkely in 1965.

Fuzzy Linguistic Variables are used to represent qualities spanning a particular spectrum. Fuzzy Control combines the use of fuzzy linguistic variables with fuzzy logic. It is represented based on the inputs, outputs, and Disjunction and Conjunctions within specific rules. And it includes several basic steps such as fuzzification: to Calculate Input Membership Levels and fuzzification: to Constructing the Output and Find centroids (when the Location where membership is 100%).

Fuzzy logic is one of machine learning algorithms, which is used in the case of relative data, so that the values are not specified as the terms that humans use [26], so the computer cannot deal with it, as if the data are vague, such as cold, hot, very cold, wet, and very humid, for such values are not considered clear and cannot be determined, so the data are divided into fuzzy sets, for example, it can be described as less crowded, very crowded or moderate crowding. All of them are not specified, but it can be determined by making the category of 5-10 people less crowded and from 30-60 very crowded, in case the area of the place is small and so on.

*F. Building the Fuzzy Model*

In this section, the crowding of worshipers in the mosque area has represented using a Linguistic Fuzzy term set of three labels (no crowd, Medium crowd, and crowd). The three Labels were divided into percentage of 0 to 100 as follows (see Fig. 3):

- 0-30% (No Crowd): In this case, the ratio of the number of worshipers is calculated with the capacity of the mosque, and if it is below 30% it is represented that there is no crowding.

- 25-75% (Medium Crowd): In this case, if the ratio of the number of worshipers to the capacity of the mosque is between 25% and 75%, then the case is medium crowding.

- 70-100% (High Crowd): In this case, if the ratio of the number of worshipers to the capacity of the mosque is more than 70% then the case is crowd.

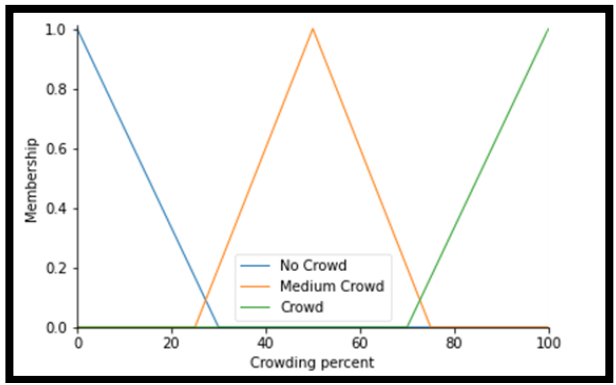

Fig. 3. Linguistic Fuzzy Term Set of Three Labels for Crowding.

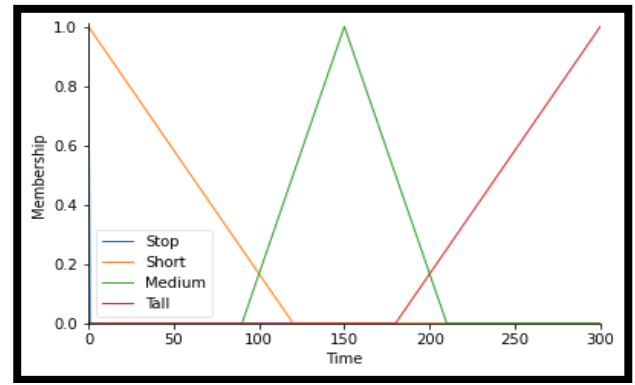

Fig. 5. Linguistic Fuzzy Term Set of Four Labels for Dome's Status.

As a result of image processing according to the above proportions, the total number of worshipers in the mosque have achieved. Then, the weather and temperature conditions have obtained to determine the appropriate cases for opening the domes. The weather conditions were divided into two labels (Rain and outlook) as follows (Fig. 4):

- (0° and 24°): represents the minimum and maximum temperatures for rainy conditions.
- (7° and 47°): represents the minimum and maximum temperatures for moderate climates.

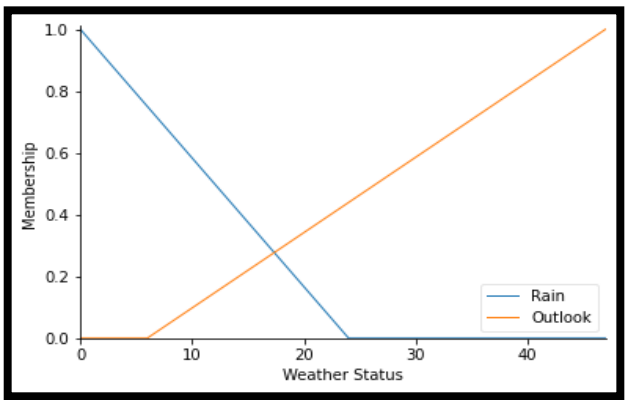

Fig. 4. Linguistic Fuzzy Term Set of Two Labels for Weather Status.

The weather result, combined with the results of image analysis, are inputs that define the state of the dome opening and the time needed per second to conduct the necessary ventilation to the mosque as follows (see Fig. 5):

- State (0): The dome remains closed, and this state working if it was raining.
- State (0-120): The dome opens for a short period, and this state working if the weather was good and there was no crowding.
- State (90-210): The dome opens for a medium period, and this state occurs if the weather was good, and crowding was medium.
- State (180-300): The dome opens for a long time, and this state occurs if the weather was good and there was high crowding.

After train the model according to the inputs from the weather data and the number of worshipers, the following rules have been generated (see Fig. 6):

- Rule 1: in this case, when only if the weather status is "Rain" then we must close the domes (time to open the domes is 0 so the result is "stop").
- Rule 2: in this case, when the weather status is "Outlook" and the status of a crowd is "no crowd" then the time to keep the domes open must be "short".
- Rule 3: in this case, when the weather status is "Outlook" and the status of a crowd is "Medium Crowd" then the time to keep the domes open must be "medium".
- Rule 4: in this case, when the weather status is "Outlook" and the status of a crowd is "Crowd" then the time to keep the domes open must be "Tall".

```
rule1 = ctrl.Rule(weather['Rain'], time['Stop'])
rule2 = ctrl.Rule(weather['Outlook'] & crowd['No Crowd'], time['Short'])
rule3 = ctrl.Rule(weather['Outlook'] & crowd['Medium Crowd'], time['Medium'])
rule4 = ctrl.Rule(weather['Outlook'] & crowd['Crowd'], time['Tall'])
```

Fig. 6. Generated Logical Rules for Dome's Status.

IV. RESULTS AND DISCUSSION

The computations involving fuzzy sets, linguistic models, and CSRnet have done using Python programming language. The weather dataset used for the experiments has obtained from Saudi Arabia's weather history on Kaggle website [20]. The detailed description of the dataset, implementation details about the linguistic terms were discussed in the previous sections.

In this section, the results will be discussed in detail. As mentioned in the earlier section, two inputs "weather status and crowding rates", and one output "dome status" have been generated. However, to achieve the aim of this work, the proposed model should be tested by combining the previous inputs with taking into account the proportions for each attribute and linking them with the output.

The following scenario has been tested assuming that the crowding rate of the mosque with the worshippers in the mosque is 72%, and the temperature recorded 30° (see Fig. 7, Fig. 8, and Fig. 9).

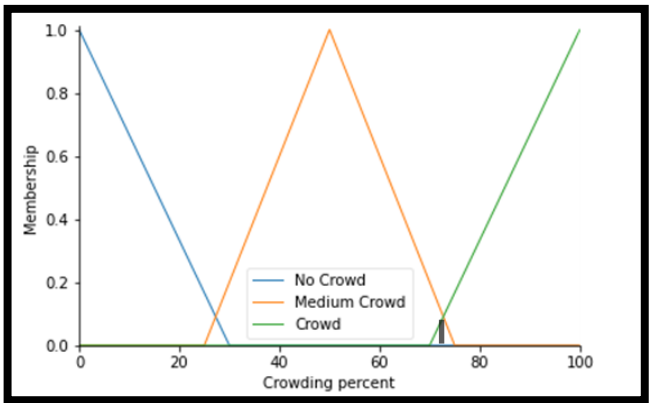

Fig. 7. Crowding Rate of the Mosque with the Worshippers is 72%.

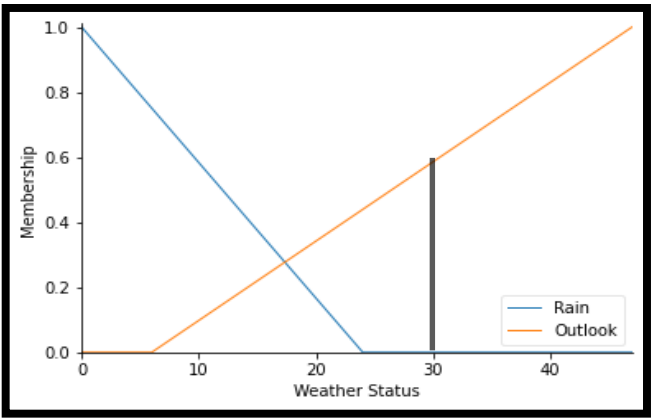

Fig. 8. Weather Status when the Temperature is 30°.

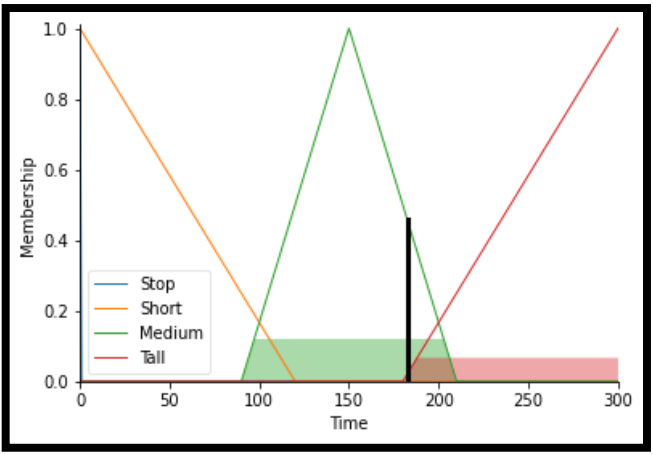

Fig. 9. Testing Scenario for the Proposed Model.

The previous Fig. 7 shows the crowding rate of the mosque with the worshippers when it is 72%, while Fig. 8 shows the weather status when the temperature is 30°. In Fig. 9 both the crowding rate and the temperature degree were combined and represented as inputs, and according to the previously generated rules (see Fig. 6), the exact period during which the domes must be open every hour head can be determined, as in the proposed scenario, this relationship intersected with both the time (medium and tall) so the relationship will be presented in the minutes and according to the inputs the domes will open for 3.04 minutes.

It can be concluded from the previous scenario that the exact time for opening and closing the domes can be specified when the crowding rate and the weather conditions are defined using Congested Scene Recognition Network (CSRNet) and Fuzzy techniques.

## V. CONCLUSION AND FUTURE WORK

Since the spread of mosques is increasing, as well as the technological development in a permanent race over time, it became necessary to coordinate the development of mosques and technological development to match each other, thus the availability of a comfortable environment during worship and approach to Allah. Today we are in the age of using latest technologies in everything, based on this principle a proposed model for the smart dome was presented as the beginning of the computerization of mosques. The Prophet's Mosque was adopted to apply this model, knowing that it can be applied to any mosque provided the necessary materials and techniques.

The proposed model has built by Congested Scene Recognition Network (CSRNet) and Fuzzy techniques using Python programming language to control the domes to be opened and closed for a specific time to renew the air inside the mosque. The proposed model has worked efficiently and specifies the exact duration time to keep the domes open for a few minutes for each hour head. This research came as an extension of previous published research [6] with more modifications and improvements to it, and promising results have been obtained in this work.

As an extension to the model proposed, more factors that affect the dome status might be added to have more accurate results to control the dome movement. Also, different machine learning Algorithms can be applied [27], [28] and deep learning.

REFERENCES

[1] Michael Lipka, "Muslims and Islam: Key findings in the U.S. and around the world", PEW Research Center/FACT TANK, August 9, 2017.

[2] Maryam Ghayyada, "How many mosques are there in the world", Available: mawdoo3.com / Home / Islamic landmarks, 1 July 2019.

[3] Muhammad bin Abdullah Al-Sabeel, "A brief summary about the architecture of the Two Holy Mosques, since the introduction of Islam to the era of the Custodian of the Two Holy Mosques", 2019.

[4] Mohammed bin Ali Al-Thobyani Al-Juhani, Sabah Saudi,"The largest expansion in the history of the Prophet's Mosque", 2020.

[5] Al-Khuzayem, Bandar bin Muhammad, Miller, & Abdul Rahman. "Estimating the crowd density in the Grand Mosque using the bypass neural network". 2019.

[6] Mohammad Awis Allababede, Anas H. Blasi, Mohammad Alsuwaiket, "Mosques Smart Domes System using Machine Learning Algorithm", International Journal of Advanced Computer Science and Applications (IJACSA) : Vol.11, No.3, 2020.

[7] Yuhong Li, Xiaofan Zhang, Deming Chen, "CSRNet: Dilated Convolutional Neural Networks for Understanding the Highly Congested Scenes", arXiv : 1802.10062v4 [cs.CV] 11 Apr 2018.

[8] LEA, Robert N., et al. "An HVAC fuzzy logic zone control system and performance results". In: Proceedings of IEEE 5th International Fuzzy Systems. IEEE, p. 2175-2180. 1996.

[9] AZAZA, M., et al. "Fuzzy decoupling control of greenhouse climate. Arabian Journal for Science and Engineering", 40.9: 2805-2812, 2015.

[10] Eldursi, S., Alamoudi, N., Haron, F., Aljarbua, F., & Albakri, G. "Crowd Density Estimation System for Al-Masjid Al-Haram". Int'l

Journal of Computing, Communications & Instrumentation Engg. (IJCCIE) Vol. 4, Issue 1, ISSN 2349-1469 EISSN 2349-1477, 2017.

[11] Rahman, S. M., Rabbi, M. F., Altwijri, O., Alqahtani, M., Sikandar, T., Abdelaziz, I. I & Sundaraj, K., "Fuzzy logic-based improved ventilation system for the pharmaceutical industry". J. Eng. Technol, 7, 640-645, 2018.

[12] GOUDA, Mohamed Mahmoud. "Fuzzy ventilation control for zone temperature and relative humidity". In: Proceedings of the American Control Conference, IEEE, p. 507-512. 2005.

[13] JOMAA, Manel, et al. "Greenhouse modeling, validation and climate control based on fuzzy logic". Engineering, Technology & Applied Science Research, Vol. 9, Issue 4, P.4405-4410, 2019.

[14] LIN, Chin-Teng, et al. "Neural-network-based fuzzy logic control and decision system". IEEE Transactions on computers, 40.12: 1320-1336. 1991.

[15] PAL, A. K.; MUDI, R. K. "Self-tuning fuzzy PI controller and its application to HVAC systems". International journal of computational cognition, 6.1: 25-30, 2008.

[16] AL-AHMADI, Hassan M., et al. "Statistical analysis of the crowd dynamics in Al-Masjid Al-Nabawi in the city of Medina, Saudi Arabia". International Journal of Crowd Science, 2018.

[17] Saudi Arabia weather history data. Available: Kaggle Website: https://www.kaggle.com/esraamadi/saudi-arabia-weatherhistory .Accessed on: Dec. 10, 2020.

[18] Shenoy, Arun P., AMEER, P. M. "Anamoly Detection in Wireless Sensor Networks". Conference (TENCON). IEEE. p. 1504-1508. 2019.

[19] ADHIWIBOWO, Whisnumurti; DARU, April Firman; HIRZAN, Alauddin Maulana. "Temperature and Humidity Monitoring Using DHT22 Sensor and Cayenne", API. Jurnal Transformatika, 17.2: 209-214, 2020.

[20] Malhotra, M., Aulakh, I. K., Kaur, N., & Aulakh, N. S., "Air Pollution Monitoring Through Arduino Uno". In ICT Systems and Sustainability (pp. 235-243). Springer, Singapore. 2020.

[21] Shanghai Technology Database. Available: Kaggle Website: https://www.kaggle.com/tthien/shanghaitech? Accessed on: Dec. 5, 2020.

[22] Anas H. Blasi, "Performance increment of high school students using ANN model and SA algorithm". Journal of Theoretical and Applied Information Technology 95(11):2417-2425. 2017.

[23] Rawabi A Aroud, Anas H. Blasi, Mohammed Alsuwaiket. "Intelligent Risk Alarm for Asthma Patients using Artificial Neural Networks". International Journal of Advanced Computer Science and Applications, Vol. 11 No. 3, 95-100, 2020.

[24] Anas H. Blasi, Mohammed A. Alsuwaiket. "Analysis of Students' Misconducts in Higher Education Institutions using Decision Tree and ANNs". Engineering, Technology and Applied Science Research. Vol. 10 (No. 6): 6510-6514. 2021.

[25] Zadeh, Lotfi A., and R. A. Live. "Fuzzy Logic Theory and Applications", World Scientific Publishing Company, 2018.

[26] Anas H. Blasi, "Scheduling Food Industry System using Fuzzy Logic," Journal of Theoretical and Applied Information Technology, vol. 96, no. 19, pp. 6463–6473, Oct. 2018.

[27] Mohammed A. Alsuwaiket, Anas H. Blasi, khawla Altarawneh. "Refining Student Marks based on Enrolled Modules' Assessment Methods using Data Mining Techniques". Engineering, Technology and Applied Science Research. Vol. 10(No. 1):5205-5010. 2020.

[28] Anas H. Blasi, Mohammad A. Abbadi, Rufaydah Al-Huweimel. "Machine Learning Approach for an Automatic Irrigation System in Southern Jordan Valley". Engineering, Technology and Applied Science Research. Vol. 11 (No. 1): 6609-6613. 2021.